# Synthesis of supervised classification algorithm using intelligent and statistical tools

Supervised Classification

Ali DOUIK
U.R.: Automatique Traitement de Signal et Image (ATSI)
Ecole Nationale d'Ingénieurs de Monastir
Monastir – TUNISIE
Ali.douik@enim.rnu.tn

Mourad MOUSSA JLASSI
U.R.: Automatique Traitement de Signal et Image (ATSI)
Ecole Nationale d'Ingénieurs de Monastir
Monastir – TUNISIE
Mourad.enim@yahoo.fr

*Abstract*— A fundamental task in detecting foreground objects in both static and dynamic scenes is to take the best choice of color system representation and the efficient technique for background modeling. We propose in this paper a non-parametric algorithm dedicated to segment and to detect objects in color images issued from a football sports meeting. Indeed segmentation by pixel concern many applications and revealed how the method is robust to detect objects, even in presence of strong shadows and highlights.

In the other hand to refine their playing strategy such as in football, handball, volley ball, Rugby..., the coach need to have a maximum of technical-tactics information about the on-going of the game and the players. We propose in this paper a range of algorithms allowing the resolution of many problems appearing in the automated process of team identification, where each player is affected to his corresponding team relying on visual data. The developed system was tested on a match of the Tunisian national competition. This work is prominent for many next computer vision studies as it's detailed in this study.

## I. INTRODUCTION

In the last ten years, motion detection and analysis have become very important for a wide range of applications, especially since complex algorithms can nowadays be processed real-time. Examples of applications that use motion segmentation techniques are gesture recognition, tracking applications [1, 2, 3, 4, 5], video surveillance systems [6, 7], industry, robotics [14], the medical field [15], aeronautics [17], Pattern Recognition [13] and recently, sports sector [18]. Although a lot of research has done in this field on objects segmentation, still a lot of difficulties have to be considered in this area, especially to produce good results in changing circumstances. The main purpose of this paper assignment is to present an overview of objects segmentation techniques and classification.

A large variety technique has developed and improved, K. Karman et al. used Kalman filter to model a dynamic background. Similarly K. Elgammal et al. [9] presented a non-parametric background model to model dynamic background. Toyama et. al. [10] used Wiener filter to make a linear prediction of the pixel intensity values, given the pixel historic. C. Wren et. al. [11], use a single Gaussian model per pixel and the parameters are updated by alpha blending. Unfortunately, these approaches fail in case the distribution of the background colour values do not fit into a single model. Ying Ming et al. [12] worked out a statistical algorithm inspired from the idea of Elgammel based on Cauchy distribution; they proved that ratios of intensity values between the background pixels and the current image pixels are adapted to Cauchy's distribution. In fact it is characterized by a little wide form covering the tails of the histogram; on the other hand Gaussian distribution has an exponential form.

Several works was done concerning classification, Pal et al. [21] proposed an SVM technique, their work reports the results of two experiments in which multi-class SVMs are compared with Maximum Likelihood (ML) and Artificial Neural Network (ANN) methods in terms of classification accuracy; SVM achieves a higher level of classification accuracy than either the ML or the ANN classifier.

Classification by artificial vision in soccer sector has been largely mediatized and became a significant research topic. The result of a match has serious consequences on the club life and its external environment (media, sponsors...). To refine their play strategy [20], coach and the leaders need to have technical-tactics and relevant information [19] about events of the play as well as of the players. Indeed the use of the color in computer vision application is yet very recent, musical field [22], metals classification [23], road scenes analysis and sensing domain [16, 29, 8]. In this paper various supervised classification techniques were applied. They are based on intelligent tools as fuzzy and neuronal classification on the one hand, statistic and hybrid classification based respectively on moments difference and determination of three significant color components on the other hand. A comparative study about player recognition rates was elaborated enabling us to

Ali DOUIK et al /International Journal on Computer Science and Engineering Vol.1(2), 2009, 89-97

conceive an adequate method for football players classification at the aim to classify each player in his suitable class automatically.

## II. SEGMENTATION TECHNIQUES EASE OF USE

### A. Detection by histogram analysis

In artificial vision field, colour images are taken by a video camera and then digitized by a computer. Since the soccer video is taken by static camera, the supporter and useless information can be removed by delimiting the playing zone with an affine function. Objects identification in colour images is a relevant stage in classification study; therefore we begin by the background subtraction to detect players and then to classify each of them in its suitable class. The algorithm is based on the following stages:

- Convert the original image to standards rgb levels removing the light reflections.
- Detect the high and low thresholds of each histogram then carry out the threshold on the three chromatic levels.
- Apply a logical operator "AND" on the three levels and the original images.

### B. Detection by statistical learning

Detection by histogram analysis consists to segment colour images and to remove useless information that have no contribution in classification phase. This method isn't a good choice of segmentation for many applications as presented in figure 2. The major problem of this technique corresponds at the time when the background and foreground have the same characteristics, hence after thresholding histograms many false detection can be occurred: for example it can remove a large part of useful information hence we should develop an appropriate technique.

Because the parametric background model still lacks flexibility when dealing with non-static backgrounds, a highly flexible non-parametric technique is proposed for estimating background probabilities from many recent samples over time using Kernel density estimation.

In the non-parametric model all recently observed pixel values $x_1, x_2 \ldots x_N$ are modeled by probability density functions using a certain kernel estimator function, which is often chosen to be a Gaussian. The weighted sum of all these Gaussians results in the final probability density function of the pixel value $x_t$:

$$p(x_t) = \frac{1}{n} \sum_{i=1}^{N} K(x_t - x_i) \quad (1)$$

The kernel estimator function K is chosen to be a Normal function $N(0, \Sigma)$ with $\Sigma$ being the kernel function bandwidth. For simplicity reasons the color channels within $\Sigma$ are assumed independent, but each with their own kernel bandwidth $\sigma_j^2$. Because of these assumptions the final density estimation can then be written as:

$$\Pr(x_t) = \frac{1}{K} \sum_{i=1}^{k} \prod_{j=1}^{d} \frac{1}{\sqrt{2\pi\sigma_j^2}} e^{-\frac{(x_{tj} - x_{ij})^2}{2\sigma_j^2}} \quad (2)$$

When this probability is below a certain threshold, the pixel is classified as a foreground pixel. The threshold can be adjusted to achieve a desired proportion of false positives. In other kernel density estimation applications, the kernel bandwidth has to be dependent on the number of samples. When there are many samples the bandwidth should be smaller than if we have few one. However, in this case temporal properties are taken into account for determining the $\sigma_j^2$ of color channel j. For each color channel the median m of the absolute differences of each consecutive pairs of samples is calculated. We estimate $\sigma$ by:

$$\sigma = \frac{m}{0,68\sqrt{2}} \quad (3)$$

This method guaranties that the local deviation is large when there are many large jumps between consecutive samples and smaller when this is not the case.

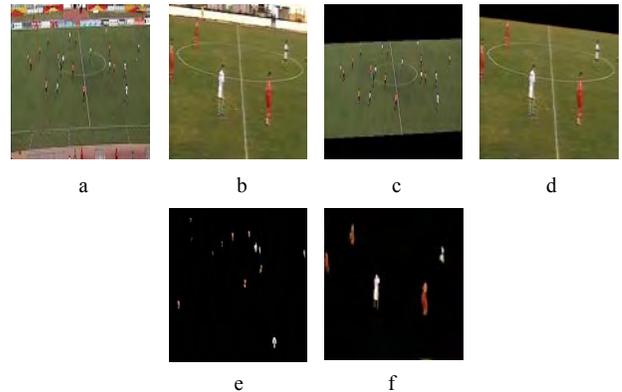

Figure 1. (a, b) Represent original images, (c, d) Represent delimited images, (e, f) Represent segmented images.

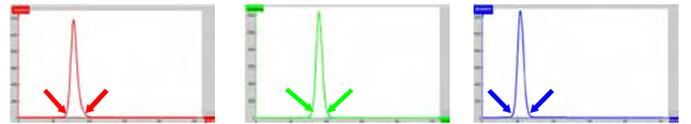

Figure 2. Histograms of standards RGB levels (r, g, b)

### C. Singular Value Decomposition Approach [24]

#### 1) Introduction

A non-parametric background modeling technique, has been applied on a soccer video images, the main problem that can be appears is the occurring of wrong detection pixels, indeed shadow pixels are detected as moving objects result to an over segmentation that will be damage, many later works where this paper it's registered, however this algorithm is extremely important because it's a part of players classification and tracking on a soccer video.

Two major issues in the SVD technique: (1) carrying out a mathematical approach and (2) explain main advantage of the method proposed here and showing influences of the singular values choice on a treated output image, besides we will see the prominent contribution using SVD theory to restore and eliminate shadow, highlights and noise from camera displacement and changed circumstances.

*2) SVD approximation of an image*

The main objective of background segmentation technique is to use singular value decomposition of a given image A represented by a matrix $A_p = [a_{ij}]$, when it can be decomposed into a product of three matrix $U_k S_k V_k^T$ as shown in figure 3. Where $a_{ij}$ is the appearance frequency of chromaticity and intensity of background pixels (p= red, green, blue).

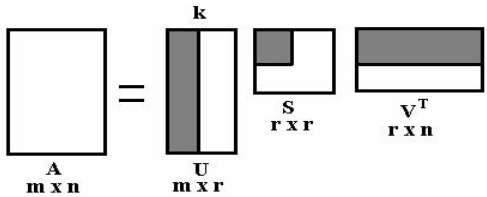

Figure 3.  Singular values decomposition of the Matrix A.

SVD technique consists to reduce the size for each initial chromatic level from r to k rank by suppression of r-k column. Matrix S represents diagonally the singular values classified by decreasing orders. Low values have no influence on the total energy of A. $U_k$ and $V_k$ are orthogonal matrices issued from matrices U and V. The determined singular values for each plan were presented in frequency space. There representation proves that for each one corresponds a discrete frequency. The noise that can occur in the signal (in frequency space the amplitude of noise is constant) corresponds to a low amplitude of singular value whereas high amplitudes of these represents global signal energy.

*3) Confidence intervals research*

In this section, we describe the basic background model and the background subtraction process with singular value decomposition. It's both used in the restoration or the reconstruction of an image, to increase the compactness distribution of different classes and also to provide useful image information.

To evaluate mathematical contribution of SVD, a quantification of global signal energy distribution according to the weight of each singular value $S_{kk}$ was done. The figure 4 illustrates the energy distribution E defined by:

$$E = \sum_{i=1}^{k} A_k^2 \quad (4)$$

The relative energy contained by each singular value K, noted $p_k$ is defined by:

$$p_k = \frac{S_{kk}^2}{E} \quad (5)$$

Where the energy of the K singular value is equal to $S_{kk}^2$.

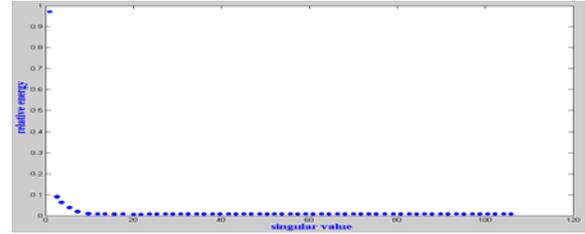

Figure 4.  Weight distribution showing 7 dominant representing 99 % of the signal energy among 108 weights

As it's shown in figures 5a, 5b and 5c, the size of treated image will be deduced from the curves representing standard déviation of each colour levels according to the singular value decomposition. In fact a good choice of the size leads to reduce both compactness in different distributions and in computing time. According to figures 5a, 5b and 5c, we can denote two zones, the first one defined in the interval $\left[0, (S_{kkl})_i\right]$ where $(S_{kkl})_i$ is the singular value limits corresponding to the linear part of the curve (i = red, blue, green), in this zone the curve presents a slope, beyond $(S_{kkl})_i$ a second zone appears where the standard deviation varied slightly therefore the optimal singular value $(\hat{S}_{kkl})_i$ must necessarily belong to the first zone of each curve.

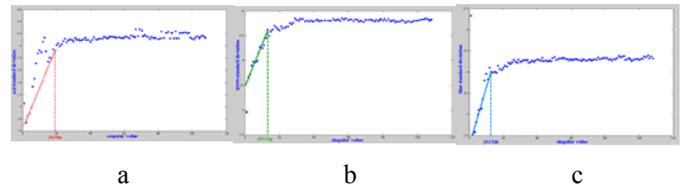

a        b        c

Figure 5.  Evaluation of standard deviation according to singular values respectively the red, green and blue channel.

Table1 illustrates initial and improved standard deviations for three channels (RGB).

TABLE I.     EVALUATION OF IMPROVEMENT PARAMETERS

| | | | |
|---|---|---|---|
| **R** | 4.2044 | 119.61 | 1156.9 |
| **G** | 4.7227 | 152.08 | 1073.3 |
| **B** | 4.313 | 88.988 | 1141.7 |
| **R_SVD** | 3.9274 | 119.07 | 1229.8 |
| **G_SVD** | 4.6204 | 152.05 | 1104.2 |
| **B_SVD** | 3.9954 | 88.407 | 1205.5 |

The choice of singular values will be kept depending on two issues: the first one is the energy curve evaluated by figure 4 and the second one is the standard deviation curves of each chromatic level shown in figures 5a, 5b and 5c. In fact we specify for each component the singular value limit previewed. The table 2 shows confidence intervals as well as the limit and the optimal singular values that vary from a level to another.

TABLE II. SPECIFICATION OF CONFIDENCE INTERVALS

|  | $(S_{kkl})$ | $(\hat{S}_{kk})$ | confidence intervals |
|---|---|---|---|
| **Red plan** | 29 | 19 | [0, 29] |
| **Green plan** | 28 | 13 | [0, 28] |
| **Blue plan** | 19 | 13 | [0, 19] |

*4) Foreground segmentation and shadow suppression*

Using the probability $\Pr(x_t)$ calculated in equation 2, the pixel is considered as a foreground pixel if $\Pr(x_t) < th$. The threshold th is a global threshold over all the image that can be adjusted to achieve a desired percentage of false positives. The shadows detection as foreground regions is a source of confusion for subsequent phases of analysis. Color information [18] is useful for shadows suppression by separating color from lightness information. For a given three color variables, R, G and B, the chromaticity coordinates r, g and b(r=R/(R+G+B), g=G/(R+G+B), S=(R+G+B)/3). Consider the case where the background is completely static, and let the expected value for a pixel be $(r_i, g_i, s_i)$. Assume that this pixel is covered by shadow and let $(r_t, g_t, s_t)$ be the observed value for this pixel at this frame. Then it is expected that $th_1 < s_t / s_i < th_2$.

To prove robustness of this algorithm, the figure 6 illustrates different player windows and shows how we can overcome segmentation (removing shadow pixels and pixels background having the same characteristics that those foreground pixels).

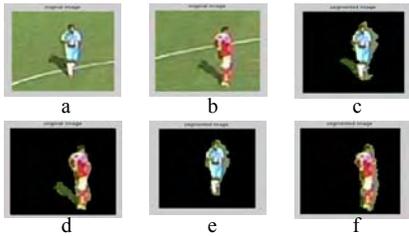

Figure 6. Effect of the singular value decomposition level. Where (a and b) are Original images, (c and d) Detection using chromaticity coordinates r, g and the lightness variable s. (e and f) Detection using chromaticity coordinates r, g and s with SVD.

### III. CLASSIFICATION ALGORITHMS

A large variety of supervised classification algorithms was developed, ranged from statistical to intelligent tools and they operate on only color regions. Each one of them is expressed in adapted color system representation that will contribute to an optimal classification.

*A. Hybrid classification*

The advantage of this algorithm is that the color will be represented in a system of the three most discriminating levels, in order to be able to separate the colour nuance distributions corresponding to pixel players from each team.

*1) Hybrid colour system*

After extraction of useful information which represents the pixel players, we separate the two classes using colorimetric analysis. Nevertheless traditional RGB space cannot be the most discriminating representation space. Indeed, other colour systems, deduced from the RGB components, can be more suitable according to the considered case. For this reason, treatment of pixel players in various colour systems leads to a hybrid space represented by the three best colour components.

*2) Method description*

N.Vandenbroucke [29] considered a multidimensional space composed of the chromatic levels currently used as the following:

E = {R, V, B, r, v, b, X, Y, Z, I1, I2, I3, y, i, q, u, v, l, t, s}. In each level α (α∈E) the algorithm of discrimination is based on various phases:

- Select three training player windows $J_{1A}$, $J_{2A}$ and $J_{1B}$ in the RGB system.
- Convert in the α level player windows.
- Calculate the average of pixel coordinates (x, y) representing each player:

α (x, y): pixel value (x, y) in the plan α.

$R_{1A}$ and S ($R_{1A}$) are respectively area and surface of the player $J_{1A}$.

$R_{2A}$ and S ($R_{2A}$) are respectively area and surface of the player $J_{2A}$.

$R_{1B}$ and S ($R_{1B}$) are respectively area and surface of the player $J_{1B}$.

Therefore we can evaluate for each level α:

α-average region for $J_{1A}$: $\quad m^{\alpha}_{1A} = \dfrac{\sum_{R_{1A}} \alpha(x, y)}{S(R_{1A})}$

α-average region for $J_{2A}$: $\quad m^{\alpha}_{2A} = \dfrac{\sum_{R_{2A}} \alpha(x, y)}{S(R_{2A})}$

α-average region for $J_{1B}$: $\quad m^{\alpha}_{1B} = \dfrac{\sum_{R_{1B}} \alpha(x, y)}{S(R_{1B})}$

- Calculate the distance between $J_{1A}$ and $J_{2A}$ as well as the distance between $J_{1A}$ and $J_{1B}$

$$D^{\alpha}_{1A,2A} = \left| m_{1A}^{\alpha} - m_{2A}^{\alpha} \right| \quad (6)$$

$$D^{\alpha}_{1A,1B} = \left| m_{1A}^{\alpha} - m_{1B}^{\alpha} \right| \quad (7)$$

Where these expression $D^{\alpha}_{1A,1B}$ and $D^{\alpha}_{1A,2A}$ are respectively the distance between $J_{1A}, J_{2A}$ and $J_{1A}, J_{1B}$ in α level. The discriminating criterion adopted is determined as the difference $D^{\alpha}$ between these two distances:

$$D^{\alpha} = D^{\alpha}_{1A,1B} - D^{\alpha}_{1A,2A} \quad (8)$$

We proceed with the same way for all α level from E set. The criterion value Classification in a descending order lead to the determination of the most discriminating chromatic components (table 3).

TABLE III. EVALUATION OF DISCRIMINATE LEVELS

|  | α | β | γ |
|---|---|---|---|
| **Hybrid color space** | Saturation | Green | Blue |

*3) Region modelling*

After the conversion of RGB image for the players A and B in the hybrid system, the modelling phase consists in assigning to each P(x, y) of a level the average value of the pixels intensities for the corresponding level [30].

*4) Parameters and methodology*

The attributes used in this algorithm are the average values of each colorimetric component whereas the criteria of decision making are the Euclidian distance between A (s, v, b) and JA where A the coordinates for model A and JA current player to be classified (table 4)

. Similarly we calculate the distances separating the model B from JB.

TABLE IV. EUCLIDIAN DISTANCE BETWEEN TWO OBJECTS AND MODELS

| d2 (B, J$_A$) | 0,54616 | d2 (A, J$_A$) | 0,063875 |
|---|---|---|---|
| d2 (B, J$_B$) | 0,022593 | d2 (A, J$_B$) | 0,63185 |

The membership's decision of each player belonging to a team is specified by evaluating the distance that separates player window and the two models. Indeed, a very weak distance corresponds to players of the same cluster; the opposite case corresponds to two players of different cluster.

The results of classification by this method are illustrated in tables 5 with a very encouraging classification rate of around 93 %.

TABLE V. CLASSIFICATION RATE FOR THE HYBRID TECHNIQUE

| Classe | Total number of players | Number of classified players | Percentage of classification |
|---|---|---|---|
| Team A | 210 | 201 | 95 % |
| Team B | 210 | 195 | 92 % |

*B. Difference moments algorithm (dmom)*

The colour moments are measurements that can be employed to differentiate images based on their colors characteristics. They provide a measurement of the similarity between color images. The parameter values of each model can be compared with the images constituting database. The colour moments [26] can be considered as a discriminating criterion between texture and color objects. In fact the colour distribution for these regions follows a certain density of probability (Gaussian).

*1) Used parameters*

In this method Sticker and Orengo [27] employed three central colour moments expressed as follows:

- The average: $\quad E_i = \frac{1}{N} \sum_{j=1}^{N} P_{ij}$

Where $E_i$ is the average value of the player pixels to be classified

- Standard deviation: $\sigma_i = \sqrt{(\frac{1}{N} \sum_{j=1}^{N} (P_{ij} - E_i)^2)}$

$\sigma_i$ represents the dispersion degree between pixels player and their average.

- Skewness: $\quad S_i = \sqrt[3]{(\frac{1}{N} \sum_{j=1}^{N} (P_{ij} - E_i)^3)}$

Si represents the asymmetry degree between pixels player and their average.

*2) Methodology*

This method is carried out in HSV system (Hue, Saturation and Value). The moments of candidate image that are fixed to nine are evaluated for each level. The discriminating criterion between two images (reference and candidate image) is defined as the sum of differences between the moments distributions with a weighting factor expressed by the following equation:

$$dmom(JA, MB) = \sum_{j=1}^{N} W_{i1} \left| E_i^A - E_i^B \right|$$
$$+ W_{i2} \left| \sigma_i^A - \sigma_i^B \right| + W_{i3} \left| S_i^A - S_i^B \right| \quad (9)$$

dmom: represents the distance between two players.

$M_A$: Model for Player $J_A$.

$w_i$: are the weights to be specified which are related to the specific case, they can be granted so that various preferences are given to various attributes of an image. They can be modified to increase or decrease the importance of a colorimetric component which appears to be interesting. Classification by this method is carried out through several stages as follows:

**Stage 1:** Convert image RGB to HSV.

**Stage 2:** Calculate the moments matrices of the two models ($M_A$, $M_B$).

$$M_A = \begin{bmatrix} 0{,}41 & 0{,}36 & 0{,}35 \\ 0{,}22 & 0{,}16 & 0{,}19 \\ -1{,}25 & -1{,}19 & -1{,}19 \end{bmatrix} \quad M_B = \begin{bmatrix} 0.76 & 0.70 & 0.74 \\ 0.10 & 0.08 & 0.10 \\ -0.01 & -0.01 & -0.01 \end{bmatrix}$$

**Stage 3:** Calculate the moments matrix of to classify each player.

$$MA = \begin{bmatrix} 0.71 & 0.66 & 0.70 \\ 0.08 & 0.07 & 0.88 \\ -0.31 & -0.28 & -0.28 \end{bmatrix} \quad MB = \begin{bmatrix} 0.40 & 0.33 & 0.32 \\ 0.25 & 0.17 & 0.20 \\ -0.84 & -0.74 & -0.74 \end{bmatrix}$$

Moments Matrix of Player ($J_A$)  Moments Matrix of Player ($J_B$)

**Stage 4:** Calculate the *dmom* for the following matrix of weights:

$$W = \begin{pmatrix} 1 & 2 & 1 \\ 1 & 2 & 1 \\ 1 & 2 & 1 \end{pmatrix}$$

We calculate *dmom* between two models and a given object (player). It is well noticed that these distances to an attribute can contribute to make decision on classification.

TABLE VI. THE $D_{MOM}$ BETWEEN TWO OBJECTS AND MODELS

| *dmom* (B, JA) | 7,0757 | *dmom* (B, JB) | 1,7987 |
|---|---|---|---|
| *dmom* (A, JA) | 1,8814 | *dmom* (A, JB) | 5,3967 |

*3) Experimental results*

The classification by moments difference technique for two clusters are elaborated, the results are illustrated in table 7, the global classification rate for two classes reaches 78 %.

TABLE VII. CLASSIFICATION RATE FOR THE *DMOM* TECHNIQUE

| Class | The total number of players | Number of classified players | Percentage of classification |
|---|---|---|---|
| Team A | 210 | 129 | 62 % |
| Team B | 210 | 200 | 95 % |

*C. Fuzzy classification [31]*

Fuzzy logic is frequently used in computer vision, it may affect various applications. The most common among these are regulation, control and classification. Many fuzzy systems can be used in this context: Sugeno model, Tsukamoto model and Mamdani model (used in this algorithm).

*1) Statistical study*

With an extensible colour images database taken during the warming up time (at the beginning of the match), a statistical study [16] was elaborated allowing to establish correlation between parameters during classification. Figure 7a, 7b and 7c represents respectively the distribution curves of the intensities average values of the green (green_moy), the blue (blue_moy) and the red (red_moy).

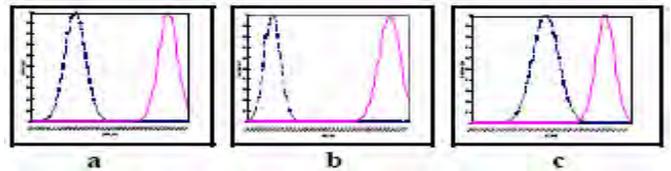

Figure 7. The intensities distribution of the three components (RGB) for two clusters

The distribution curves highlight the importance of each parameter during classification. Only the parameter red_moy presents an overlapping (Fig.7c), the two other parameters green_moy, blue_moy (Fig.7a and Fig.7b) haven't it, thus a statistical classification [25] can be used in this case.

*2) Methodology*

The fuzzy classification is done relying on the following stages:

- Fuzzification: this stage is used to quantify the input and output variables (Fig.8). This stage consists to define the membership functions of the linguistic variables.

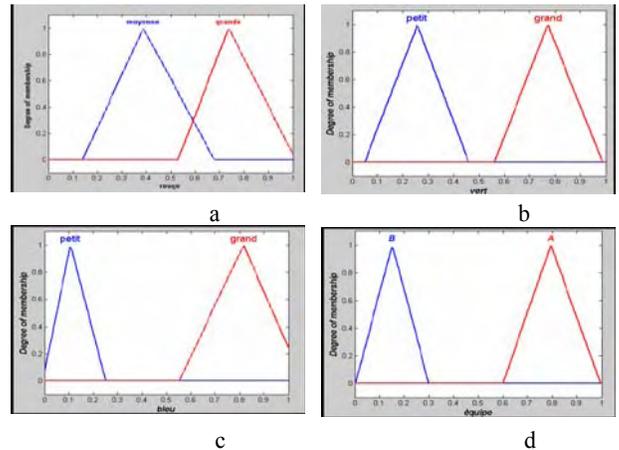

a  b

c  d

Figure 8. Membership Functions of the input and output parameters

- Inference rules: is a set of rules between the fuzzy subsets at the aim to draw deductions.

- Defuzzification: specifies the suitable output model according the inference rules, in fact the decision was made by calculating the gravity centre of the resulting function as illustrated in the following equation:

$$Zs = \frac{\int \mu s(z) z \, dz}{\int \mu s(z) \, dz} \qquad (10)$$

*3) Experimental results of classification*

The classification by fuzzy technique gives the results illustrated in table 8. The global classification rate for the two classes reaches 97 %.

TABLE VIII. CLASSIFICATION RATE FOR THE FUZZY TECHNIQUE

| Class | The total number of players | Number of classified players | Percentage of classification |
|---|---|---|---|
| Team A | 210 | 206 | 98 % |
| Team B | 210 | 203 | 96 % |

D. *Neural Networks classification algorithm (NN) [32]*

*1) Introduction*

Classification by artificial vision [28] in sports field became a significant research topic; it treated several problems that can result in an overlapping between various classes. The neural network is an information processing system inspired from cells organization of the human brain. In a neural network as represented in figure 9, we distinguish three types of neurons:

- Input neurons: they have the property to gather data whose source is apart from the network.

- Output neurons: define the output layer of the network. It contains as many neurons as the number of classes to be discriminated.

- Hidden neurons: they don't have any relation with the external world, it is used to coordinate between the input and output neurons.

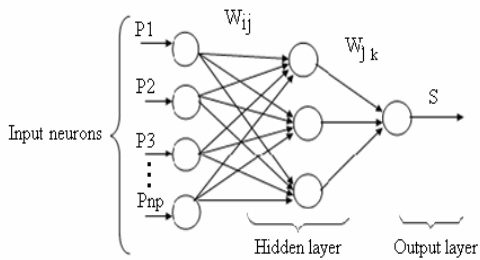

Figure 9. Neural Networks architecture

Similarly to human brain, the artificial neural networks can be learned by experiment: Indeed, the objective of supervised retropropagation algorithm is to minimize a cost function 'E' representing a quadratic error for an input-output. Where $d_K$ the desired output for the $K^{th}$ neuron and $S_K$ the obtained output by the network [20].

$$E = \sum_k (d_k - S_k)^2 \qquad (11)$$

The weight is obtained according equation (12) where α is a positive real that specifies the step of weights modification.

$$W_{ij}^{t+1} = W_{ij}^t + \alpha . C_i . S_j \qquad (12)$$

To carry out classification, we highlight relations between objects on the one hand, objects and their parameters on the other hand. In this section we present a classification by Neural Network (NN) learned with retropropagation algorithm. This technique allows discriminating clusters (players) present in color images issued from sports meeting. It can be summarized in 4 stages shown in figure10.

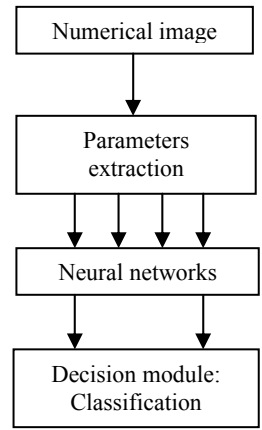

Figure 10. Regions classification Chain

*2) Training phase*

The used parameters to classify player windows by this technique are the same used in dmom technique. For the training phase 9 parameters are used, the initial base was constituted by 560 regions in various positions among this base 300 represent two clusters C1 and C2, 260 represent respectively the C3 and C4 clusters. Indeed, NN allowed weights adjustment by retro-propagation algorithm until reaching a null error defined in last section. The training phase was done as follows: we selected several player windows from the training database, and extracted parameters will be used to classify these regions. The player windows contains respectively four classes C1, C2, C3 and C4 then the desired output are indexed respectively by 1, 2, 3 and 4. To appreciate the classification algorithm, the curves representing training error according epochs number on one hand in figure 11 and error according neurons of the hidden layer on the other hand has been evaluated in Figure 12.

The choice of the neurons number in the hidden layer depends on the training phase. Indeed, we must learn networks some models in many circumstance and various positions, each experience contains a different neurons number from hidden units, the choice taken is the first one leading to a weak training error. Figure 12 shows variation of training error according to neurons number in hidden layer.

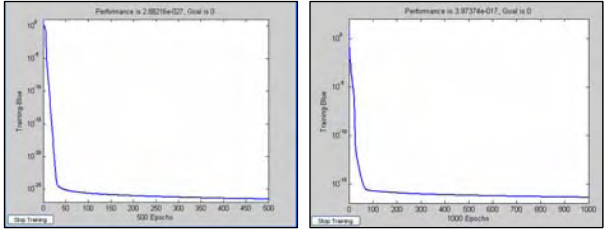

Figure 11. Evaluation of training error according to epochs number

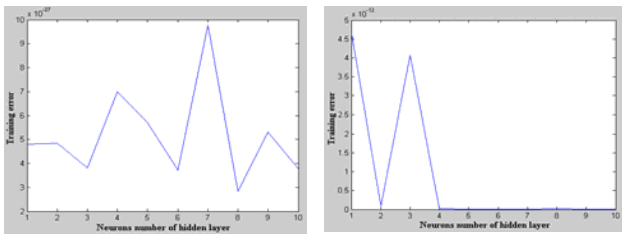

Figure 12. Evalution of training error according to neurons number in hidden layer

*3) Validation phase*

In this stage we worked with a large amount of samples from our data base with the same color classification parameters. We made the simulation of the network that is already created in the training stage. In the validation network stage we evaluated by testing 546 players of the various classes and for the two sports meeting dividing it as follows: 161, 167, 114 and 104 player windows respectively for C1, C2, C3 and C4 clusters. Figure 13 represents many colour regions detected by the segmentation algorithm. This test allows appreciating the performances of the neuronal system. If the performances are not satisfactory, it will be necessary either to modify the network architecture, or to modify the training base. The obtained classification results are illustrated in table 9.

Another test was carried out by widening the training data base by increasing it from 130 to 180 player windows for each class of second meeting and from 150 to 200 for the two other classes, this test leads to improve the recognition rate in Table 10, indeed it reaches 100 % for the most classes (C1, C3, C4).

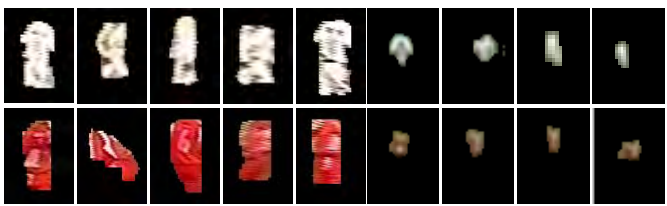

Figure 13. Player windows for each cluster for many positions

TABLE IX. CLASSIFICATION RATE FOR THE (NN) TECHNIQUE IN THE 1ST TEST

| Classe | The total number of players | Number of classified players | Percentage of classification |
|---|---|---|---|
| C 1 | 161 | 140 | 87 % |
| C 2 | 167 | 135 | 81 % |
| C 3 | 114 | 100 | 88 % |
| C 4 | 104 | 89 | 86 % |

TABLE X. CLASSIFICATION RATE FOR THE NN TECHNIQUE IN THE 2ND TEST

| Class | The total number of players | Number of classified players | Percentage of classification |
|---|---|---|---|
| C 1 | 111 | 111 | 100 % |
| C 2 | 117 | 114 | 97 % |
| C 3 | 64 | 64 | 100 % |
| C 4 | 54 | 54 | 100 % |

IV. CONCLUSION

In this paper, many algorithms were elaborated to detect objects in colour images sequences issued from sports meeting. Several automatic classification systems were made to classify different color region representing football players. All classification techniques developed are supervised, we can discriminate on the one hand intelligent techniques such as neuronal and fuzzy algorithms and on the other hand hybrid algorithm based on the determination of a color representation system containing the three most significant components and using a metric distance in this base as a decision making approach taking in to account suitable established models. In fact, we can verify in this paper that RGB system is not always the suitable system for solving several problems raised by the researchers in computer vision field (e.g. overlapping, screening…). To overcome these problems, we determined three discriminating components (v, B, S) which are not from the same representation system and leads to a promising classification rates for various color regions. Indeed, we obtained with the intelligent tools good rates, 97% and 100% respectively for fuzzy and neuronal algorithms, while for hybrid technique it reaches a rate of 98%.